\documentclass{article}

\usepackage{microtype}
\usepackage{graphicx}
\usepackage{subfigure}
\usepackage{booktabs} 
\usepackage{amsfonts} 

\usepackage{amsmath}  
\usepackage{nicefrac} 
\usepackage{url}
\usepackage{multirow} 

\usepackage{hyperref}

\usepackage[accepted]{icml2018}

\icmltitlerunning{Predicting Adversarial Examples with High Confidence}

\begin{document}
\twocolumn[
\icmltitle{Predicting Adversarial Examples with High Confidence}





\begin{icmlauthorlist}
\icmlauthor{Angus Galloway}{uog,vector}
\icmlauthor{Graham W. Taylor}{uog,vector,cifar}
\icmlauthor{Medhat Moussa}{uog}
\end{icmlauthorlist}

\icmlaffiliation{uog}{School of Engineering, University of Guelph, Canada}
\icmlaffiliation{cifar}{Canadian Institute for Advanced Research}
\icmlaffiliation{vector}{Vector Institute for Artificial Intelligence, Canada}

\icmlcorrespondingauthor{Angus Galloway}{gallowaa@uoguelph.ca}

\icmlkeywords{Machine Learning, ICML}

\vskip 0.3in
]



\printAffiliationsAndNotice{}  

\begin{abstract}

It has been suggested that~\emph{adversarial examples} cause deep learning
models to make incorrect predictions with high confidence. In this work, we 
take the opposite stance: an overly confident~\emph{model} is more likely to be 
vulnerable to adversarial examples. This work is one of the most proactive 
approaches taken to date, as we link robustness with~\emph{non-calibrated} model
confidence on noisy images, providing a data-augmentation-free path forward.
The~\emph{adversarial examples phenomenon} is most easily explained by the
trend of increasing non-regularized model capacity, while the diversity and
number of samples in common datasets has remained flat. Test accuracy has
incorrectly been associated with true~\emph{generalization performance},
ignoring that training and test splits are often extremely similar in terms of
the overall representation space. The transferability property of
adversarial examples was previously used as evidence against overfitting
arguments, a perceived random effect, but overfitting is not always
random.



\end{abstract}

\section{Introduction}
\label{introduction}

Practically obtainable datasets are inherently sparse in high-dimensions.
This is true for image classification tasks like the CIFAR-10 and ImageNet
datasets, on which deep neural networks have achieved very low test error.
The~\emph{adversarial examples} phenomenon was an observation that these same
state-of-the-art deep learning models are easily
fooled by images that are objectively very similar
to the naturally occurring data on which they were 
trained~\cite{szegedy2013intriguing}.
The effect implies two seemingly contradictory statements,
which~\citet{Jo17} and~\citet{Dube18} summarize well: on one hand, deep neural 
networks generalize extremely well to a held-out test set, yet any randomly 
selected correctly classified image is arbitrarily close to a misclassified 
one.

Several hypotheses have been proposed regarding
this phenomenon, such as the idea that adversarial examples
occupy small low-probability ``pockets'' in the manifold,
yet are dense, similar to rational numbers on the real
line~\cite{szegedy2013intriguing}. However, it would be unusual
that neural networks are learning a decision boundary anything like the
distribution of rational and irrational numbers on the number line.
Furthermore, an effect where images initialized with random noise are classified
with very high confidence~\cite{NguyenYC15} suggests that there exists at least
one class of ``non-examples'' that do not occur in low-probability pockets,
but can be found almost everywhere in the representation space not spanned by
the sparse training data.

The linearity hypothesis of~\citet{Goodfellow_explaining_adversarial}
suggests that for a model parameterized by weights $w$ with an average
magnitude of $m$, and input $x \in \mathbb{R}^n$, one need only perturb $x$
by a vector of small constants $\epsilon$ aligned in the direction of $w$ to
induce a swing in activation of $\epsilon m n$. They argue that $\epsilon$
shrinks with increasing $n$; however~\citet{TanayG16} remind us that the
magnitude of the activations also grows linearly with $n$. They demonstrate
that linear behaviour alone is insufficient to cause adversarial examples, and
our experiments confirm this.

Non-examples occupy space within the representation ability of our 
models that has yet to be explored. In high dimensions, these areas will never 
be explored, therefore it is logical to attempt to use strong regularization
to reduce representation ability such that we minimize the space away from the
training data sub-manifolds.

We show that not all methods of minimizing the unexplored space are equivalent.
Experiments on the synthetic spheres dataset 
from~\citet{gilmer_adversarial_2018} suggests that using low-precision 
representations appears to confer additional pose-invariance characteristics 
upon a model, with the added benefit of compression and easing model
deployment on general purpose hardware. In this regard, we expand on the work
of~\citet{galloway2018attacking} that compared the robustness of full-precision
and binarized models, finding that the lower precision variant was equally or
more robust to a variety of attacks.

One of our main contributions is a comparison of regularization effects of
arbitrarily low-precision internal representations against traditional methods 
of regularizing a model, such as weight decay. We then explore a fundamental 
trade-off between preserving sensitivity to valid natural image classes, and 
reducing total unexplored space. To the best of our knowledge, we are also the 
first to defend against~\emph{fooling images}~\cite{NguyenYC15}, also known as
``rubbish class'' examples, without using RBF 
networks which do not generalize well~\cite{Goodfellow_explaining_adversarial}.

\section{Background}

Regularization was investigated as a potential solution to the adversarial
problem as early as in~\citet{szegedy2013intriguing}, but was discarded
in subsequent work~\cite{Goodfellow_explaining_adversarial} after modest 
amounts of $L_1$ weight decay did not completely resolve the problem. We 
additionally suspect the transferability property: 
that adversarial examples generated on one architecture 
(e.g.~ResNets) are likely to be misclassified by others trained independently 
(e.g.~VGG, Inception), was seen as evidence~\emph{against} 
overfitting. As overfitting is traditionally viewed as a~\emph{random} effect 
that leads to poor generalization on the test set, researchers falsely 
concluded that adversarial examples and overfitting are unrelated. We maintain 
that training over-parameterized models on the same datasets, using the same
optimization procedure, is sufficient to cause non-random overfitting.

Additional work similar to regularization includes~\citet{FawziFF15},
who found that training on random noise also was not beneficial. 
Eventually, the
suggestions of~\citet{szegedy2013intriguing} were implemented
by~\citet{cisse17a}, in which the Lipschitz constant of various layers is
constrained to be $\leq 1$. The idea was to prevent instabilities from
propagating through the network, but in practice resulted in modest gains
beyond those conferred by adversarial training data-augmentation approaches.
Additionally,~\citet{cisse17a} did not conduct an in-depth analysis against
traditional regularization techniques like weight decay, beyond mentioning
that they used some weight decay in their experiments. As we show, the
``state-of-the-art'' architectures they used are very fragile to strong weight
decay, which likely discouraged them and others from exploring this further.

\citet{Gu15} proposed using ``deep contractive networks'', which they argued
makes the model more flat near the training data by introducing a smoothness
penalty inspired by a contractive autoencoder (CAE). The main idea was that
making the model's decision boundary more flat near the training data manifold
should maximize the $L_2$ distortion required to cause misclassification.
They also noticed that denoising autoencoders are able to recover 90\% of
adversarial errors by reconstructing examples as a pre-processing step.
Nonetheless, when they attempted to stack the autoencoder with a (poorly
regularized) classifier, the overall model was easily defeated in an end-to-end 
attack. Ultimately, while we do agree that smoothness is desirable
near the training data to a limited extent since an $n$-sphere maximizes 
distance to the surface for a fixed volume sub-manifold, \citeauthor{Gu15} 
do not address the behaviour of their model globally,~\emph{away} from the
training data. This likely leaves it vulnerable to the gradient ascent 
``fooling images'' attack~\cite{NguyenYC15}. Additionally, their results 
are reported in terms of the $L_2$ input distortion that causes a 100\% 
misclassification rate, which is difficult to compare with other literature on 
neural network defenses.

\citet{netdissect2017} propose a framework for automatically quantifying
disentangled representations in deep CNNs by network dissection. Although they
do not explicitly focus on robustness to adversarial perturbations, their work
is complementary to ours in that they observe a significant degree of
variability in the interpretability of different models that all obtain
very similar ``generalization'' performance on the test set.




As of ICLR 2018, the only non-certified defense that has lived up to
claims made for a white-box threat model, after a recent informal investigation
by~\citet{athalye_obfuscated_2018}, is that of~\citet{madry2018towards}.
Although the work of~\citet{athalye_obfuscated_2018} has not yet undergone
official peer review, we believe their methodology to be sound and in keeping
with techniques that defeated similar previously published defenses
with~\emph{gradient masking} tendencies~\cite{PapernotMJFCS15, Gu15}.
Many of the defeated ICLR 2018 defenses showed signs of gradient masking, or 
tested on attacks that were too weak~\cite{dhillon2018stochastic, 
buckman2018thermometer, guo2018countering, samangouei2018defensegan, 
song2018pixeldefend, xie2018mitigating}. 
Defense strategies that rely solely on gradient masking do not improve the 
richness of features learned, or affect the underlying geometry of the decision 
boundary in a meaningful way, which ultimately determines real-world robustness 
in the practical black-box setting. The white-box model, however, is convenient 
to study as it is strictly more difficult to defend against than black-box. 
Therefore, if a model is robust against the former, it also is against the 
latter.

Expectation over Transformation (EOT) is a sampling technique that
overcomes defenses relying on stochasticity by attacking them end-to-end after
averaging gradients over several (e.g.~10 or more) forward passes, before
taking a backward step. It has been shown to create reliable adversarial
examples against a variety of random viewpoint transformations, such as over 
changes in scale and rotation in 2 or 3D~\cite{athalye_synthesizing_2017}.
It is conceivable that variants of the same attack can be used against even the
most creative defenses, such as Defense-GAN, where the non-trust worthy
input image is first encoded into a latent space, substituted with its nearest
neighbor, and then classified~\cite{samangouei2018defensegan}. The attack exploits
the concurrently demonstrated notion that adversarial examples can exist
directly~\emph{on} the data manifold, for a synthetic concentric spheres data
set~\cite{gilmer_adversarial_2018}.

It is also highly plausible that non-differentiable input transformations blocking
gradient flow can confidently be back-propagated through using the same
straight through gradient estimator (STE) commonly used to train binarized and
low-precision neural networks to high accuracy~\cite{BengioLC13,
CourbariauxB16, zhou2016dorefa}. This is essentially the ``Backward Pass
Differentiable Approximation'' used to bypass defenses that destroy
gradient signal in~\citet{athalye_obfuscated_2018}.

\section{The~\citet{madry2018towards} Defense}
\label{attack}

The~\citet{madry2018towards} defense uses projected gradient descent 
(PGD)~\cite{Kurakin17a} with a random initialization to defend against 
perturbations $\mathcal{S} \subseteq \mathbb{R}^d$ allowed under the threat
model. More formally, it consists of a min-max optimization game defined 
by~\eqref{eq:minmax}. The inner maximization consists of adding a random 
(e.g.~uniform) $L_p$ bounded perturbation $\delta\in \mathcal{S}$ to the input 
$x$, and then taking several steps in the direction of the gradient that 
maximizes the training loss $J$ with respect to $x$. For each step in the 
inner maximization loop, the sign of the gradient is usually scaled by a 
constant $\epsilon$ and accumulated in $x$. The outer minimization is a normal 
update to model parameters $\theta$ by stochastic gradient descent on the 
loss obtained with this new batch of adversarial examples.
\begin{equation}
\label{eq:minmax}
\min_{\theta} \bigg( \mathbb{E}_{(x,y)\sim\mathcal{D}}\left[\max_{\delta\in
	\mathcal{S}}J(\theta,x+\delta,y)\right] \bigg)
\end{equation}
Their particular claim of robustness considers perturbations up to $\epsilon = 0.3$
and $\nicefrac{8}{255}$ under the
$L_\infty$ norm for the MNIST and CIFAR-10 data sets respectively. The
$L_\infty$ norm is a reasonable choice since it strictly maximizes all
preceeding $L_p$-norms (e.g.~$L_1, L_2$), which explains
why they see favourable performance against $L_2$ bounded attacks in their
Figure 6, despite never training with an $L_2$
bounded adversary. We can derive an upper limit for an $L_\infty$ trained model
in terms of $L_2$ robustness on an $n$ dimensional data set given that $\|x\|_2
\leq \sqrt{n}\|x\|_\infty$. This implies a theoretical upper $L_2$ limit of 
$8.4$ for
MNIST, and $\nicefrac{443}{255}$ for CIFAR-10. Interestingly, both MNIST and
CIFAR-10 models in Figure 6 of the~\citet{madry2018towards} paper use roughly 
10\% of this theoretical limit, a quantity determined by picking an accuracy 
level on the $L_\infty$ plot and mapping it to $\epsilon$ where the same value 
occurs on the $L_2$ plot. This could be explained by the choice of sampling the 
initial random perturbation from a uniform distribution, which will only have a 
few values that saturate the limits of $\pm \epsilon$. In any case, the initial 
random perturbation has proven to be critically important for increasing the 
diversity of examples trained against during training, as opposed to the 
original deterministic fast gradient
method~\cite{Goodfellow_explaining_adversarial}.

Despite the success of data-augmentation approaches to date, we share the
opinion of~\citet{Jo17} regarding their long-term suitability, as it is hard to 
guarantee that a~\emph{different} attack will not be successful against the 
particular adversary used during training. Additionally, the choice of threat 
model assumed in data-augmentation based defenses is often arbitrary, and 
performance has been found to rapidly decay for perturbations just slightly 
beyond that seen by the model in training~\cite{Kurakin17a, madry2018towards}.

\section{Experiments on the Madry Defense}

We believe it is imperative that the community move away from solely
data-augmentation based approaches as a defense mechanism. For example, the 
min-max game-theoretic nature of~\eqref{eq:minmax} assumes an attacker is 
playing the same game. This is confirmed by our experiments against 
the~\citet{madry2018towards} defense, as we find at least one~\emph{zero} 
and~\emph{first}-order ``blindspot'', where behaviour is demonstrably worse 
than un-defended or well regularized model.

\subsection{Constant Pixel Intensity Attack}
\label{sec:madry_constant_exp}

In Figure~\ref{fig:madry_mnist_sota_const_attack}, a gradient-free, or
``zero-order'', attack is deployed against the state-of-the-art models
provided online by~\citet{madry2018towards} that were secured with PGD\@.
The attack is trivial in nature, and simply involves adding a constant offset
(positive or negative) to the image, and then clipping to the valid pixel
range. There are legitimate concerns about this weakness, and many reasons
to expect the background colour to differ in a real world machine vision
application. Humans have no trouble reading digits or classifying images
registered against a clean, uniform background, even when the difference
 between the pixel intensities in the foreground and background is small.


The attack was motivated by an analysis of the learned convolutional
kernels in~\citet{madry2018towards} (in particular, Figure 9 in the paper, and
the surrounding discussion in Appendix E), in which they mention that the first
layer implements a learned thresholding filter. All but~\emph{three} of 
thirty-two kernels in the first layer of the three-layer CNN were
approximately~\emph{zero}. Among the three kernels that were not all zeros, each had 
only~\emph{one} element that was non-zero. The function of the~\emph{entire} 
first layer for the PGD trained model is to scale and threshold. 
The reason this could be an optimal strategy against a PGD adversary is that, 
in combination with learned biases, perturbations up to $\epsilon_{\max}$ are 
forced to the non-linear region of the ReLU activations.

Both of the independently trained ``secret'' and ``public'' models
converge to this configuration, and their respective values of the non-zero
elements are (1.34, 0.86, 0.60) and (1.26, 0.80, 0.52). Kernels did have a more
distributed representation in subsequent layers, but the pattern observed in 
the first layer was concerning. This is partly why we find one of the 
main recommendations made by~\citet{madry2018towards} to be problematic, namely
that ``\emph{increasing capacity improves robustness}'', given that their
models are choosing to not use the capacity they~\emph{already} have. Their
justification is that more capacity is required to train with a stronger
adversary such as PGD, but this contradicts earlier, more intuitive
findings in~\citet{TanayG16}, that a properly regularized
``nearest centroid'' classifier is unaffected by adversarial examples. We
maintain that extraneous capacity is actually a~\emph{liability} in terms of 
robustness.

Although the models are reasonably well-behaved under their limited threat 
model, we maintain that increased sensitivity to a DC offset, as shown by the 
large shift down and to the left between the cyan filled area with dash-dot 
border, to orange area, is clearly undesirable. The right edge of each filled 
area plots accuracy when the offset is subtracted from the original images, 
while the left edge is when it is added.

\newcommand{\madrywidth}{0.45} 

\begin{figure}[]
\centering{\subfigure[]{
\includegraphics[width=\madrywidth\columnwidth]{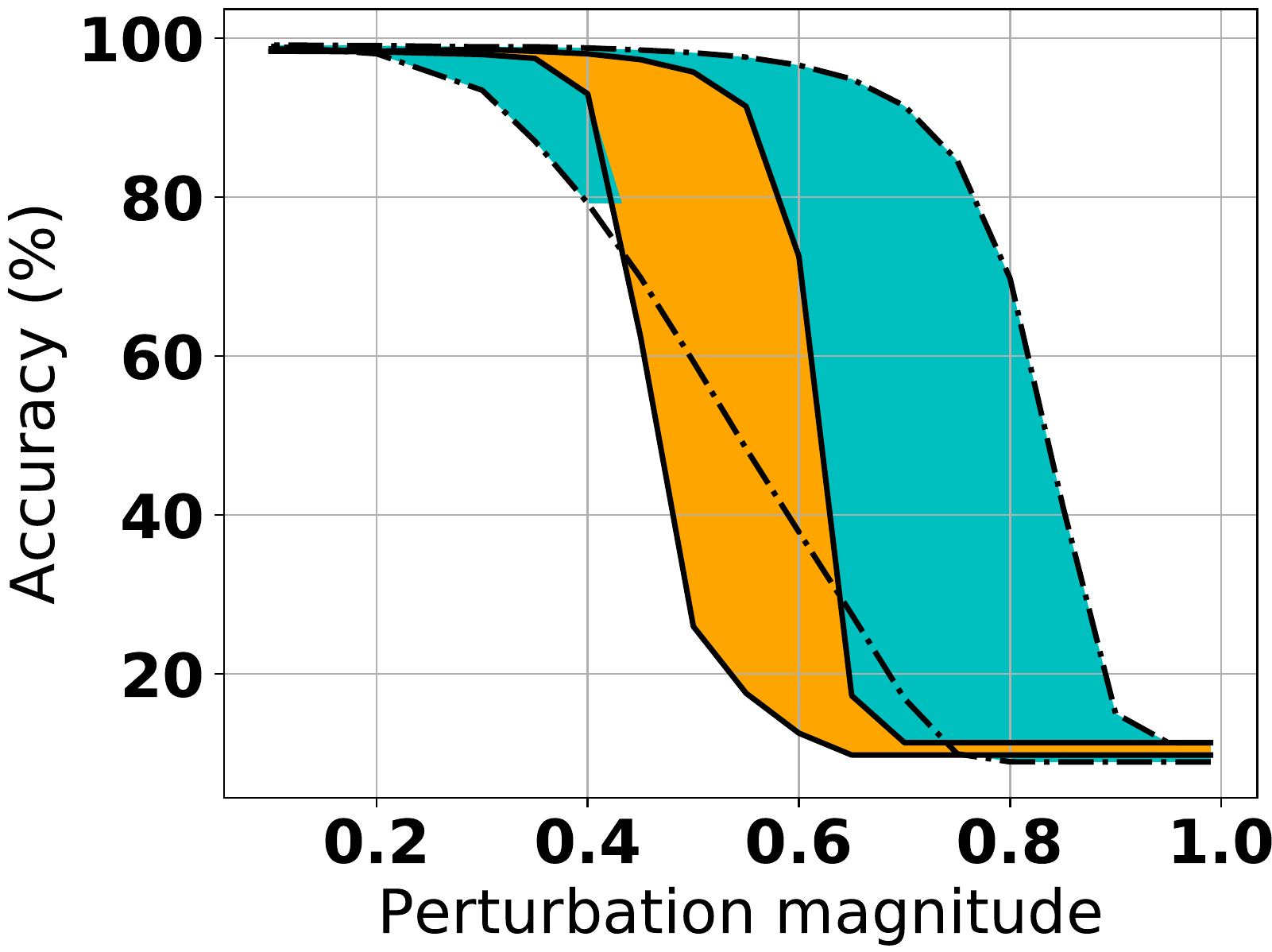}
\label{fig:madry_mnist_sota}}} 
\centering{\subfigure[]{
\includegraphics[width=\madrywidth\columnwidth]{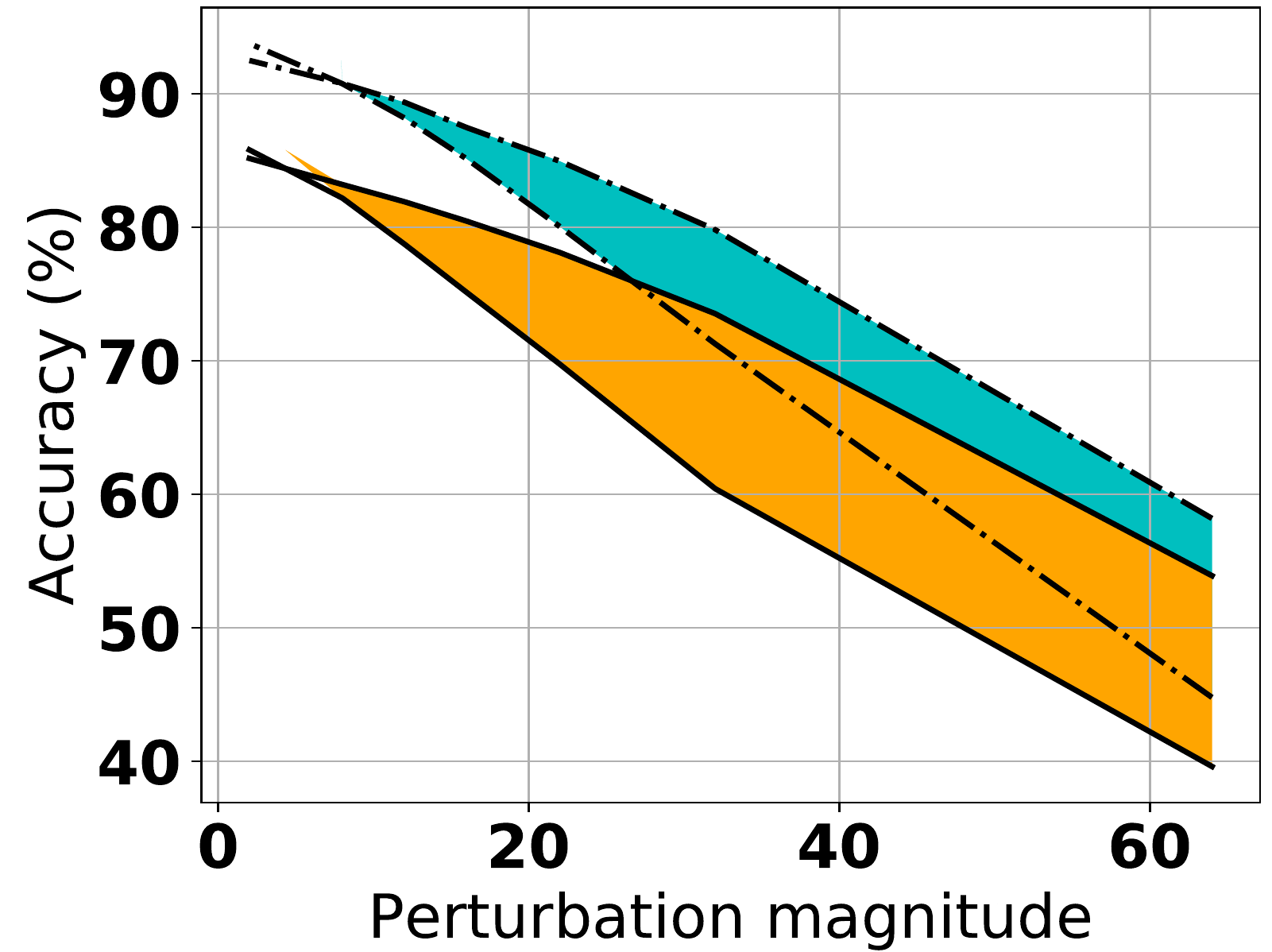}
\label{fig:madry_cifar10_sota}}} 

\caption{Attacking the state-of-the-art ``secret'' model from the black-box
~\subref{fig:madry_mnist_sota} MNIST and
~\subref{fig:madry_cifar10_sota} 
CIFAR-10 robustness challenges hosted by \citet{madry2018towards}\textsuperscript{1}.
The attack consists of adding a constant scalar value to every pixel, which can be interpreted as mostly a change in background colour. To
better characterize the two models, we test beyond the limits of their
threat model by including perturbations larger than 0.3. To do so, we
disable the check in their script that verifies the attack data set has
$\epsilon \leq 0.3$ w.r.t to the natural test set, but we still clip all
values to the valid input range.}
\label{fig:madry_mnist_sota_const_attack}
\end{figure}
\setcounter{footnote}{1}
\footnotetext{\url{https://github.com/madrylab/mnist_challenge}\\ 
\url{https://github.com/madrylab/cifar10_challenge}}

\subsection{Non-Example Gradient Ascent Attack}
\label{sec:madry_non_example_exp}

Rather than starting with correctly classified inputs and assessing how much
accuracy is maintained after perturbing them by a finite amount, we can
instead perform gradient ascent on random noise until the probability
assigned to a desired target class is maximized.
Intuitively, a~\emph{robust} model should respond in one
of two ways: the input could be classified with~\emph{low} confidence,
since it is very distant from any natural example in the training distribution.
Alternatively, if it is to be classified with high confidence, the noise should
be meaningfully transformed into something that resembles a legitimate
example. This is exactly the test that was performed by~\citet{NguyenYC15},
which found that state-of-the-art deep neural networks respond in
neither of these two ways. These models consistently classify unrecognizable 
noise images as belonging to a natural object class with very high confidence 
(e.g 99.99\%).

\newcommand{\nonexwid}{0.85}

\begin{figure}[t!]
\centering{\subfigure[Natural]{
\includegraphics[width=\nonexwid\columnwidth]{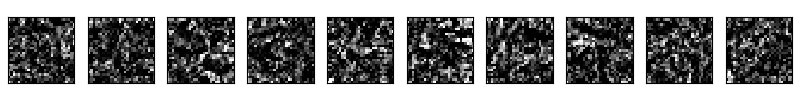} 
\label{fig:mnist_nat_grid}}} 
\centering{\subfigure[Public]{
\includegraphics[width=\nonexwid\columnwidth]{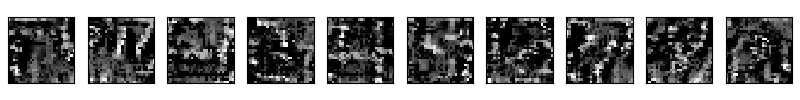} 
\label{fig:mnist_pub_grid}}} 
\centering{\subfigure[Secret]{
\includegraphics[width=\nonexwid\columnwidth]{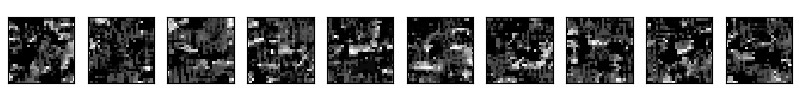} 
\label{fig:mnist_sec_grid}}} 
\caption{Non-examples classified with 100\% confidence as each of
the MNIST digits 0 through 9 (read from left to right) by the
three models from Github\textsuperscript{1}.
~\subref{fig:mnist_nat_grid} is un-protected,
while~[\subref{fig:mnist_pub_grid} and~\subref{fig:mnist_sec_grid}] are
defended with 40 iterations of PGD in the inner-loop, up to
$\epsilon_{\max}=0.3$ and step size of $0.01$. Some digits can
almost be identified, such as a ``3'', ``4'', and ``7''
in~\subref{fig:mnist_pub_grid}.}
\label{fig:madry_mnist_grid}
\end{figure}

\begin{figure}[t!]
\centering{\subfigure[Natural]{
\includegraphics[width=\nonexwid\columnwidth]{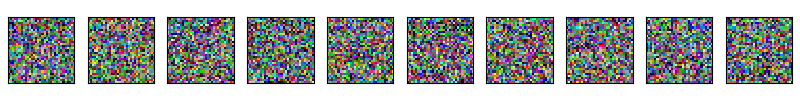} 
\label{fig:cifar10_nat_grid}}} 

\centering{\subfigure[Public]{
\includegraphics[width=\nonexwid\columnwidth]{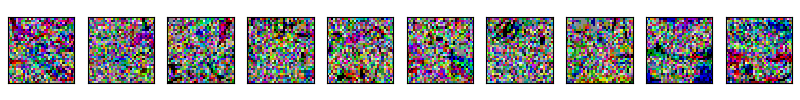} 
\label{fig:cifar10_pub_grid}}} 

\centering{\subfigure[Secret]{
\includegraphics[width=\nonexwid\columnwidth]{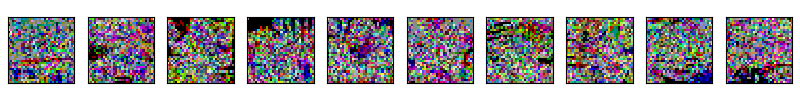} 
\label{fig:cifar10_sec_grid}}} 
\caption{Non-examples classified with 100\% confidence as each of
the CIFAR-10 classes 0 through 9 (read from left to right) by the three
state-of-the-art WideResNet (32 layers, width factor 10)
models from Github\textsuperscript{1}.
~\subref{fig:mnist_nat_grid} is unprotected, while
[\subref{fig:mnist_pub_grid} and~\subref{fig:mnist_sec_grid}] are
defended with 10 and 7 steps of PGD training respectively, with $\epsilon_{\max}=8$ and a step size of $2$.}
\label{fig:madry_cifar10_grid}
\end{figure}

The attack consists of first sampling
a noisy image $\mathcal{N}(0, 0.1) \in\mathbb{R}^{28 \times 28}$
for MNIST, and $\mathcal{N}(0, 0.1) \in \mathbb{R}^{32 \times 32 \times 3}$
for CIFAR-10. A \texttt{numpy} random seed of 0 was used,
along with $100$ steps of gradient ascent on each of the target classes using
a step size of $0.01$. All images were clipped to the valid pixel range of
[0, 1] for MNIST and [0, 255] for CIFAR-10. Identical per image standardization
is used as in~\citet{madry2018towards}. The second from right images in
Figure~\ref{fig:madry_cifar10_grid} [\subref{fig:cifar10_pub_grid}
and~\subref{fig:cifar10_sec_grid}] corresponding to the ``ship''
class have more blue pixels for water near the corners, which is
somewhat encouraging. Overall, the images are not nearly as convincing as
those in subsequent sections where strong weight decay or low-precision is used.

\section{Toy Experiments}
\label{toy-experiments}

Here, we establish intuition with toy problems that
offer significant insight into the phenomenon of adversarial examples.
We attempt to generalize observations made regarding these experiments
to challenge claims made in the literature. These experiments provide a nice lens for subsequent discussions
regarding higher-dimensional image classification problems and manifolds,
such as with CIFAR-10 and ImageNet in subsequent sections.

\subsection{Binary Classification}

We first study MNIST three vs.~seven classification, as
it is a nearly linearly separable problem for which
an optimal expert model is known a priori. By optimal, we mean a single
set of parameters, $\theta$, for a particular model architecture,
that are best suited to a particular task, and confer good robustness.
We do~\emph{not} mean that we achieve zero test error, as this
practically~\emph{guarantees} that adversarial examples will
exist~\cite{gilmer_adversarial_2018}. Zero-error implies that models
with fixed capacity are forced to cheat, by memorizing peculiarities
 or surface statistics in the data that generalize in a narrow
 distributional sense~\cite{Jo17}.

\subsubsection{Threes versus Sevens}

We wish to emphasize several points in reference to the weight
visualization from Figure~\ref{fig:mnist_three_seven}, and
empirical results from Table~\ref{tab:mnist_three_seven}.

\newcommand{\mnistwidth}{0.7in}

\begin{figure}[t]
\centering{\subfigure[]{
\includegraphics[width=\mnistwidth]{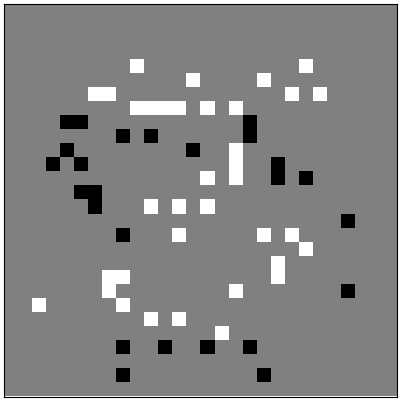}
\label{fig:lin_1b_prune}}} 
\centering{\subfigure[]{
\includegraphics[width=\mnistwidth]{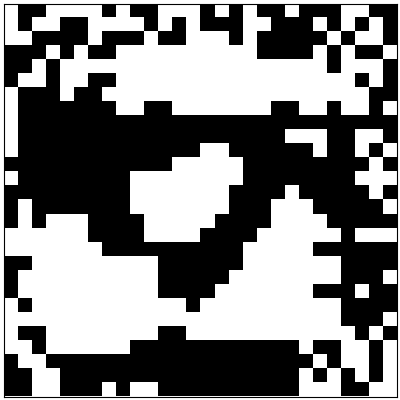}
\label{fig:lin_1b}}} 
\centering{\subfigure[]{
\includegraphics[width=\mnistwidth]{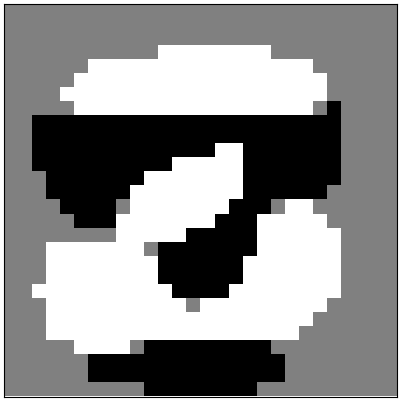}
\label{fig:lin_1p5b}}} 
\centering{\subfigure[]{
\includegraphics[width=\mnistwidth]{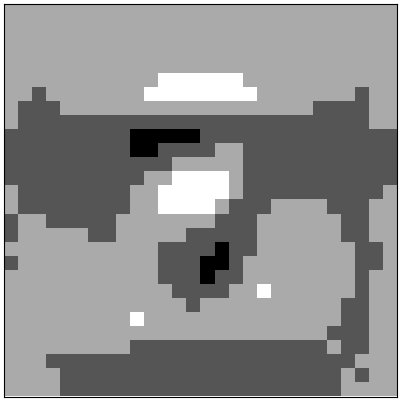}
\label{fig:lin_2b}}} 
\centering{\subfigure[]{
\includegraphics[width=\mnistwidth]{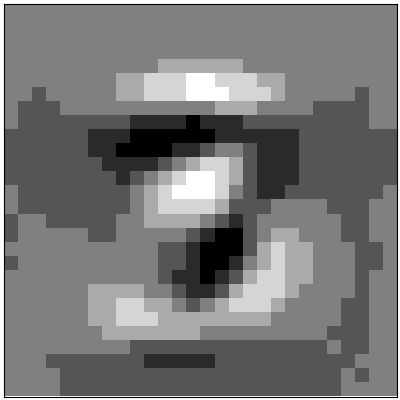}
\label{fig:lin_3b}}} 
\centering{\subfigure[]{
\includegraphics[width=\mnistwidth]{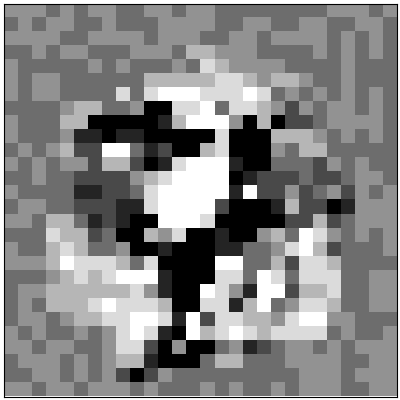}
\label{fig:lin_3b_k}}} 
\centering{\subfigure[]{
\includegraphics[width=\mnistwidth]{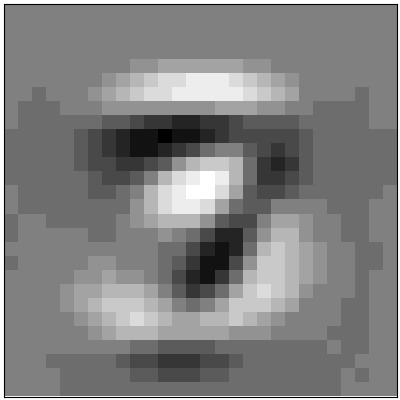}
\label{fig:lin_4b}}} 
\centering{\subfigure[]{
\includegraphics[width=\mnistwidth]{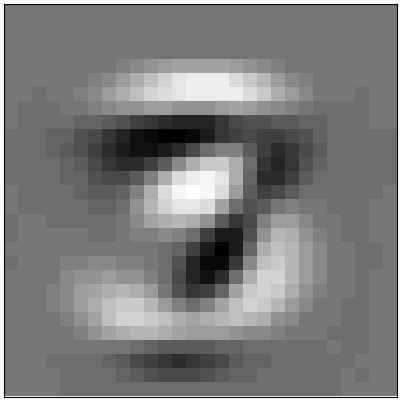}
\label{fig:lin_5b}}} 
\centering{\subfigure[]{
\includegraphics[width=\mnistwidth]{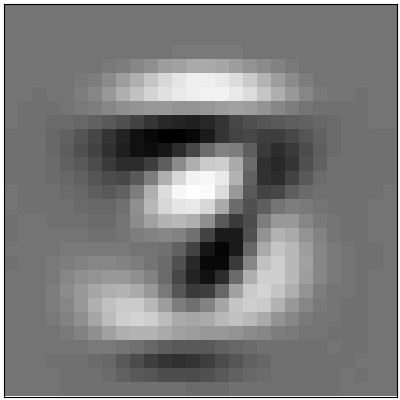}
\label{fig:lin_6b}}} 
\centering{\subfigure[]{
\includegraphics[width=\mnistwidth]{img/linear_6b}
\label{fig:lin_7b}}} 
\centering{\subfigure[]{
\includegraphics[width=\mnistwidth]{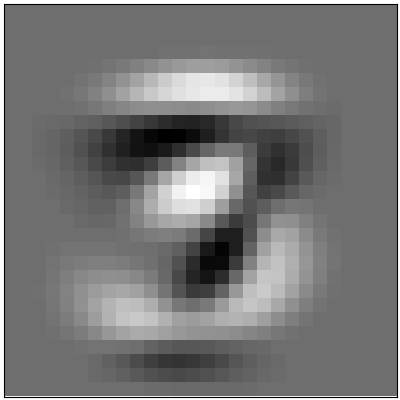}
\label{fig:lin_32b}}} 
\centering{\subfigure[]{
\includegraphics[width=\mnistwidth]{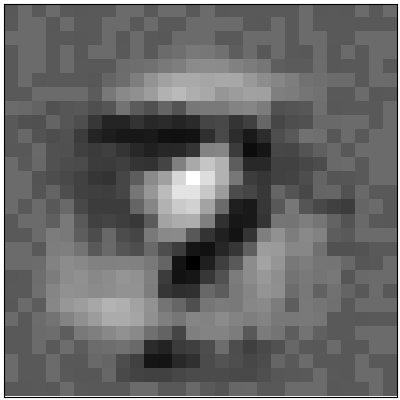}
\label{fig:lin_32b_k}}} 
\caption{Depiction of a logistic regression model's weights for
various levels of quantization after training on
an MNIST three vs.~seven binary classification task.
Models [\subref{fig:lin_1b_prune}--\subref{fig:lin_1p5b}] are 1-bit,
\subref{fig:lin_2b} is 2-bit, [\subref{fig:lin_3b} and
\subref{fig:lin_3b_k}] are 3-bit, [\subref{fig:lin_4b},
\subref{fig:lin_5b}, \subref{fig:lin_6b}, \subref{fig:lin_7b}]
are 4, 5, 6, and 7-bits respectively, while [\subref{fig:lin_32b}
and \subref{fig:lin_32b_k}] are 32-bit precision.
All weights were initialized by subtracting the
average seven from the average three, except for in
\subref{fig:lin_1b_prune} and \subref{fig:lin_1b}. 
Models were fine-tuned for 50k steps by Adam using the sigmoid cross
entropy loss, a batch size of 128, learning rate of 1e-5,
and L2 weight decay with a regularization constant of 5e-2.
Models \subref{fig:lin_3b_k} and~\subref{fig:lin_32b_k} were
trained with the same hyper-parameters, but on
adversarial examples~\cite{Kurakin17a}.}
\label{fig:mnist_three_seven}
\end{figure}

\begin{table}[]
\caption{Accuracy of the logsitic regression models from 
Figure~\ref{fig:mnist_three_seven} on test set and adversarial examples 
generated by fast gradient (sign) 
method~\cite{Goodfellow_explaining_adversarial} with $\epsilon = 0.1$. 
Models were given ``expert initialization'' except~\ref{fig:lin_1b_prune}
and~\ref{fig:lin_1b} which were initialized from a truncated normal with 
$\sigma = 0.1$.}
\label{tab:mnist_three_seven}
\vskip 0.15in
\begin{center}
\begin{small}
\begin{sc}
\begin{tabular}{ccccl}
\toprule
Model & Bits & Test & FGM & Note \\
\midrule
\ref{fig:lin_1b_prune} & 1 & 95.8 & 66.7 & Rand init + Prune \\
\ref{fig:lin_1b} & 1 & 96.7 & 34.0 & Rand init\\
\ref{fig:lin_1p5b} & 1 & 96.2 & 55.9 & Prune \\
\ref{fig:lin_2b} & 2 & 94.7 & 52.2 & \\
\ref{fig:lin_3b} & 3 & 95.8 & 73.8 & \\
\ref{fig:lin_3b_k} & 3 & \textbf{97.0} & 18.8 & Adv.~training\\
\ref{fig:lin_4b} & 4 & 95.3 & 77.3 & \\
\ref{fig:lin_5b} & 5 & 95.8 & 82.8 & \\
\ref{fig:lin_6b} & 6 & 95.8 & 84.1 & \\
\ref{fig:lin_7b} & 7 & 95.8 & 84.5 & \\
n/a	 & 8 & 95.8 & \textbf{84.8} & \\
n/a  & 32 & 94.8 & 80.1 & No training \\
\ref{fig:lin_32b} & 32 & 95.0 & 81.1 & \\
\ref{fig:lin_32b_k} & 32 & 96.7 & 82.3 & Adv.~training\\
\bottomrule
\end{tabular}
\end{sc}
\end{small}
\end{center}
\vspace{-8mm}
\end{table}

Clean test accuracy was a weak measure of model fitness. For example, 
model~\ref{fig:lin_3b_k}, which had the highest test accuracy, was also the 
most vulnerable to attack. Adversarial training does not always help. In this 
case, it was desirable to be completely invariant to certain areas of the input 
corresponding to background pixels. Naive adversarial training implementations 
actually interfere with learning this optimal solution. Additionally, a model 
that was already well-tuned hardly benefited from adversarial training 
(e.g.~compare~\ref{fig:lin_32b} and~\ref{fig:lin_32b_k}), and several 
undefended low-precision models with 5--8 bit representations outperformed the 
augmented full-precision one~\ref{fig:lin_32b_k}.

Very low-precision representations (e.g.~below 4-bits) eventually
become a compromise in terms of accuracy on both the clean and perturbed test 
sets, as the number of parameters was already very low, and  an insightful 
initialization scheme and training procedure were already 
known. Models with weights that were not given ``expert initialization''
were far more sensitive to the choice of hyper-parameters, such as the
standard deviation of the truncated normal distribution that was used to 
initialize models trained from scratch. This suggests that the robustness 
of much higher capacity models will be even more sensitive to such 
hyper-parameters.

Some configurations, such as the pruned binary model~\ref{fig:lin_1b_prune}, 
fared surprisingly well in light of the learned weight matrices. However, this 
model is likely more vulnerable to single pixel or other $L_0$ constrained 
attacks like the Jacobian Saliency Map Attack~\cite{PapernotMJFCS15}. We must 
be careful when drawing conclusions from only one type of attack. Here we only 
reported against the $L_\infty$ norm FGSM\@.

This simple experiment leads to some common sense observations about robustness,
and how it can be obtained through visualization where possible and benefitting 
from strong human judgement when available. Adversarial training or 
data-augmentation is not the only way to confer robustness, and in fact, it can
exacerbate the problem if applied to a poorly tuned model. We acknowledge that 
the fast gradient method is not the only measure of robustness, but it is 
convenient and exact for the linear model used here. We explore stronger 
iterative attacks in subsequent sections. This experiment helps us expand on 
the observation of~\citet{gilmer_adversarial_2018} that the capacity required 
to achieve~\emph{zero} error on a given dataset is significantly greater than 
that required to achieve very~\emph{low} error. In this experiment, driving a 
model with fixed architecture and capacity to zero error makes it brittle, 
therefore the definition of acceptably~\emph{low} error requires sound human 
judgement. For this architecture and dataset combination, we suggest aiming 
for no less than $\approx5\%$ error to generalize well.

\subsubsection{Spheres}

To gain further insight we reproduce the dataset and two hidden-layer MLP 
from~\citet{gilmer_adversarial_2018} in which adversarial examples can be found 
on the manifold of a synthetic $n$-sphere dataset. The task is to 
classify two n-dimensional concentric spheres with differing radii. 
We make the problem more challenging by training on a semi-sphere and testing 
on the full-sphere. We notice the 1-bit weight, 2-bit activation model retains 
a tighter shape than the full-precision equivalent, which begins to sag 
and expand where there is a lack of support.

\newcommand{\spherewid}{1.25in}
\begin{figure}[t!]
\centering{\subfigure[]{
\includegraphics[width=\spherewid]{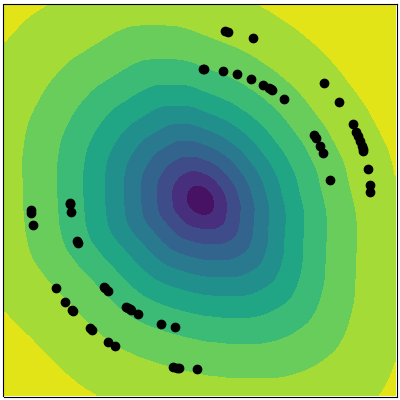}
\label{fig:spheres_32w32a}}} 
\centering{\subfigure[]{
\includegraphics[width=\spherewid]{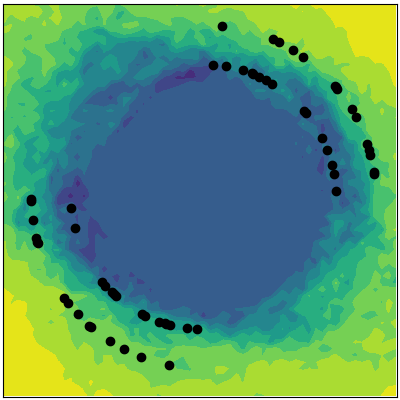}
\label{fig:spheres_1w2a}}} 
\caption{Depiction of decision boundary for MLPs trained on a variant of
the 2-dimensional Spheres dataset from~\citet{gilmer_adversarial_2018}, where
we remove two opposing quadrants from the training set, but test on a
full-sphere, a representation of higher-dimensional data sets that
are inherently sparse. The full-precision MLP~\subref{fig:spheres_32w32a}
flattens out in areas of the manifold that aren't supported, whereas the
1-bit weight 2-bit activation MLP~\subref{fig:spheres_1w2a} retains a
full-shape. Occam's razor suggests that given a lack of evidence, it is
desirable to preserve the simpler hypothesis~\subref{fig:spheres_1w2a} of
``sphere'' than a more complicated~\subref{fig:spheres_32w32a} ``ellipse''.
Best viewed in colour.}
\label{fig:spheres_2d}
\end{figure}

The lack of direct feedback between the real-valued weight update and the
thresholding done in the forward pass could be seen as a form of
inertia. One way to interpret the result in Figure~\ref{fig:spheres_2d} is that 
the binary weights vote for the status-quo and avoid contorting their decision 
boundary into unusual shapes in the absence of strong evidence.

\section{Non-Example Gradient Ascent Attack}

We now show that reducing unexplored space through strong regularization mostly 
overcomes 
this limitation, such that unrecognizable images are in fact classified 
with~\emph{low} confidence. Furthermore, when we perform gradient ascent on an 
initial noise image, we obtain plausible legitimate images for each of the 
respective classes, without using a decoder or reconstruction objective of any 
kind.

Obtaining ground-truth calibrated confidence and uncertainty estimates for
predictions made by deterministic models is
somewhat of an open problem~\cite{Gal2016Uncertainty}, but for comparison
purposes we adopt the same procedure as in~\citet{NguyenYC15}, where confidence
is measured as the magnitude of the largest softmax probability. This measure
of confidence is fairly standard in the non-adversarial examples
literature, and various ``softmax smoothing'' approaches have been
proposed to make these estimates more reliable, such as temperature
scaling~\cite{Guo2017Calibration} or penalizing high entropy 
distributions~\cite{pereyra2017regularizing}. The former approach is very
similar to the defeated defensive distillation approach that similarly 
employed softmax temperature scaling~\cite{papernot2016distillation}, providing 
a false sense of security~\cite{carlini2016defensive}.

For both Sections~\ref{sec:gen_mnist} and~\ref{sec:gen_cifar10}, we use
the following vanilla three-layer convolutional network described
by Table~\ref{tab:vanilla-cnn} with ReLU activations, and a linear
softmax readout layer. Batch normalization~\cite{batch_norm}
is only used with low-precision layers to ensure quantization is 
centered around zero.

\begin{table}[t!]
\centering
\caption{CNN architecture for experiments in Sections~\ref{sec:gen_mnist} 
and~\ref{sec:gen_cifar10}.}
\label{tab:vanilla-cnn}
\begin{tabular}{ccccccc}
\hline 
\multicolumn{1}{|c|}{Layer} & \multicolumn{1}{c|}{H} & \multicolumn{1}{c|}{W} & \multicolumn{1}{c|}{$C_{in}$}  & \multicolumn{1}{c|}{$C_{out}$}  & \multicolumn{1}{c|}{stride} & \multicolumn{1}{c|}{padding} \\ \hline 
\multicolumn{1}{|c|}{\texttt{Conv1}} & \multicolumn{1}{c|}{8} & 
\multicolumn{1}{c|}{8} & \multicolumn{1}{c|}{$img_{ch}$} & 
\multicolumn{1}{c|}{$n_f$}  & \multicolumn{1}{c|}{2}      & 
\multicolumn{1}{c|}{SAME}  \\ \hline 
\multicolumn{1}{|c|}{\texttt{Conv2}} & \multicolumn{1}{c|}{6} & 
\multicolumn{1}{c|}{6} & \multicolumn{1}{c|}{$n_f$}  & \multicolumn{1}{c|}{$2 
\times n_f$} & \multicolumn{1}{c|}{1}      & \multicolumn{1}{c|}{VALID} \\ 
\hline 
\multicolumn{1}{|c|}{\texttt{Conv3}} & \multicolumn{1}{c|}{5} & 
\multicolumn{1}{c|}{5} & \multicolumn{1}{c|}{$2 \times n_f$} & 
\multicolumn{1}{c|}{$2 \times n_f$} & \multicolumn{1}{c|}{1}      & 
\multicolumn{1}{c|}{VALID}  \\ \hline 
\end{tabular}
\end{table}

\vspace{-3mm}

\subsection{Generating MNIST examples}
\label{sec:gen_mnist}

We now attempt to explain why it is possible to generate such natural images 
without using a generative model. It was recently shown that adversarial 
examples for the MNIST dataset span an approximately 25 dimensional 
sub-manifold, which is mostly shared between models, thus enabling \emph{attack 
transferability}~\cite{Tramer17}. As we will see in comparison to ImageNet, 
this is a comparatively modest volume of unexplored example space to be
eliminated. Further, MNIST classes are relatively densely sampled in the 
training set with respect to their low dimensionality, therefore strong $L_2$ 
weight decay is sufficient to compress the representation space. Interestingly, 
the binarized models seemed to always yield plausible pen strokes independent of
other hyper-parameter settings, suggesting they could be a good starting point.

We emphasize that the images in Figure~\ref{fig:mnist_1b_gen} were generated
with an ordinary convolutional neural network, and that no decoder or
reconstruction penalty is used to modify the normal training procedure in any
way. All examples are classified with significantly less confidence than
the unrecognizable~\emph{fooling images} from~\citet{NguyenYC15}.

\newcommand{\smwidth}{0.45in}
\begin{figure}[t!]
\centering{\subfigure[]{
\includegraphics[width=\smwidth]{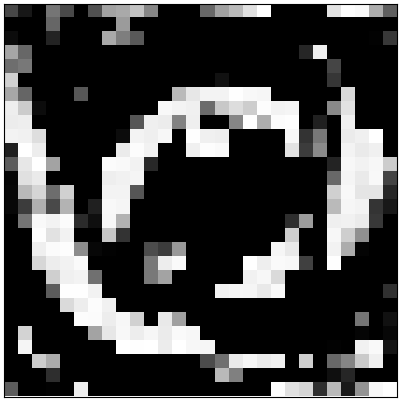}
\label{fig:mnist_1b_0}}} 
\centering{\subfigure[]{
\includegraphics[width=\smwidth]{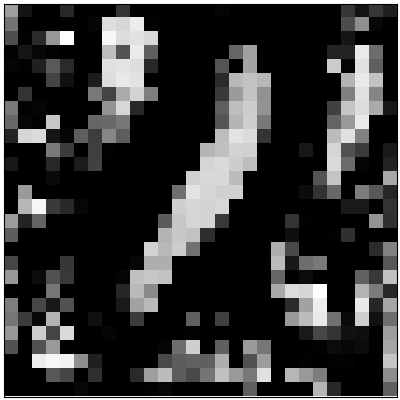}
\label{fig:mnist_1b_1}}} 
\centering{\subfigure[]{
\includegraphics[width=\smwidth]{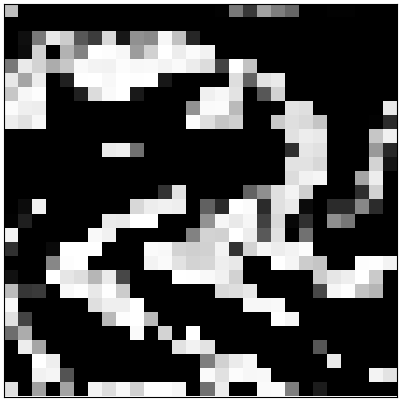}
\label{fig:mnist_1b_2}}} 
\centering{\subfigure[]{
\includegraphics[width=\smwidth]{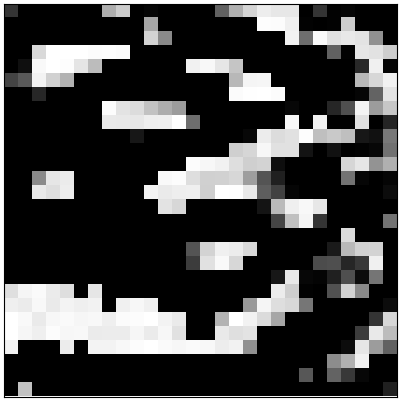}
\label{fig:mnist_1b_3}}} 
\centering{\subfigure[]{
\includegraphics[width=\smwidth]{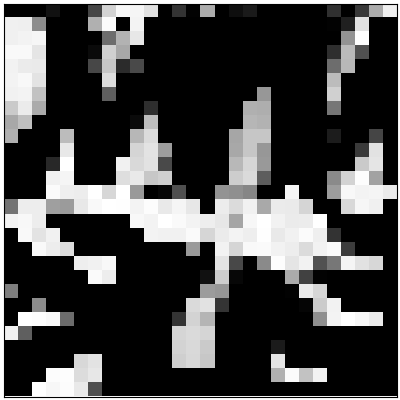}
\label{fig:mnist_1b_4}}} 
\centering{\subfigure[]{
\includegraphics[width=\smwidth]{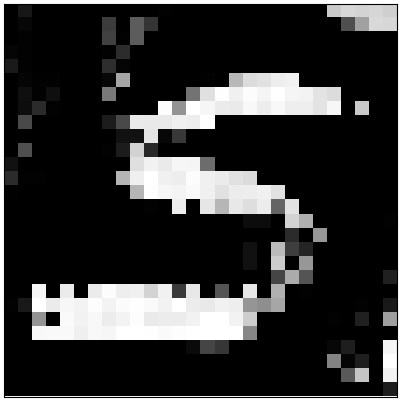}
\label{fig:mnist_1b_5}}} 
\centering{\subfigure[]{
\includegraphics[width=\smwidth]{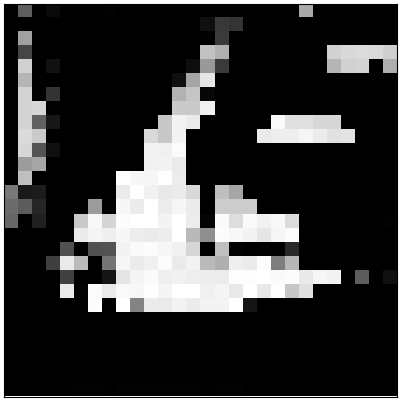}
\label{fig:mnist_1b_6}}} 
\centering{\subfigure[]{
\includegraphics[width=\smwidth]{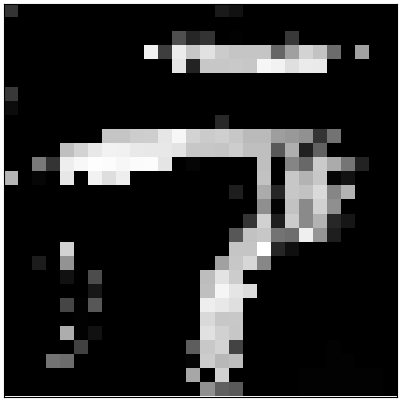}
\label{fig:mnist_1b_7}}} 
\centering{\subfigure[]{
\includegraphics[width=\smwidth]{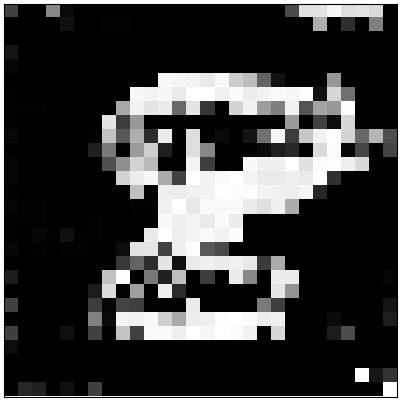}
\label{fig:mnist_1b_8}}} 
\centering{\subfigure[]{
\includegraphics[width=\smwidth]{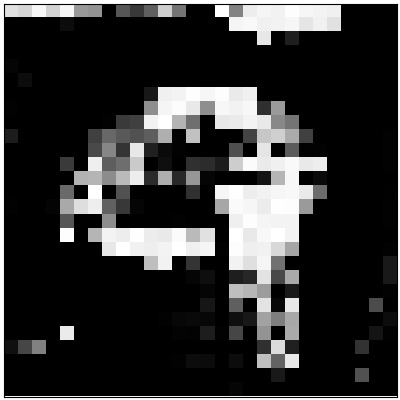}
\label{fig:mnist_1b_9}}} 
\caption{Generating MNIST examples with 1-bit (\emph{top}) and full-precision 
(\emph{bottom}) models after 100 iterations of 
gradient sign ascent on each target class with a step size of 1e-2. 
An $L_2$ regularization constant of $\lambda=0.5$ was used with the 
full-precision model and $\lambda=0.05$ on the first layer of the 1-bit model
which was retained in full-precision. Examples are classified with
45.6\%--78.0\% [\subref{fig:mnist_1b_1} and \subref{fig:mnist_1b_2}] 
for the 1-bit model, and 35.2\%--52.2\% [\subref{fig:mnist_1b_7} and 
\subref{fig:mnist_1b_8}].}
\label{fig:mnist_1b_gen}
\end{figure}

\subsection{Generating CIFAR-10 examples}
\label{sec:gen_cifar10}

In this section, we extend the same approach to the CIFAR-10 data
set~\cite{Krizhevsky09learningmultiple}. To better contextualize the images in 
Figure~\ref{fig:cifar_32b_pgd_gen}, compare with the equivalent noisy images 
that were classified with 100\% confidence by 
``state-of-the-art'' 
models from~\citet{madry2018towards} in Figure~\ref{fig:madry_cifar10_grid}.

\newcommand{\cifarwid}{0.45in}

\begin{figure}[]
\centering{\subfigure[]{
\includegraphics[width=\cifarwid]{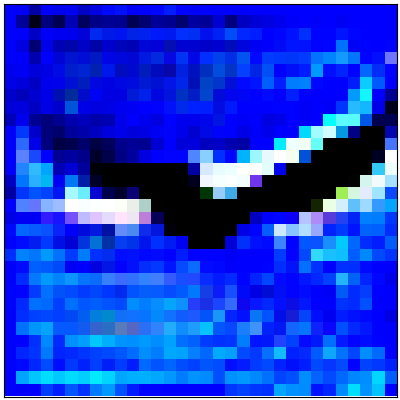}
\label{fig:cifar_32b_pgd_0}}} 
\centering{\subfigure[]{
\includegraphics[width=\cifarwid]{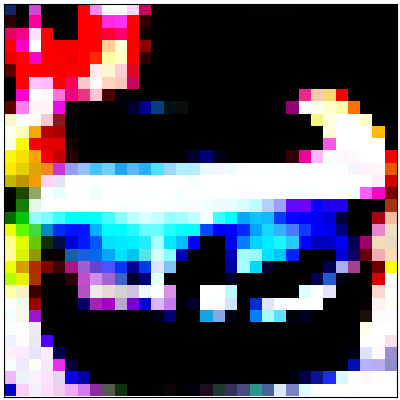}
\label{fig:cifar_32b_pgd_1}}} 
\centering{\subfigure[]{
\includegraphics[width=\cifarwid]{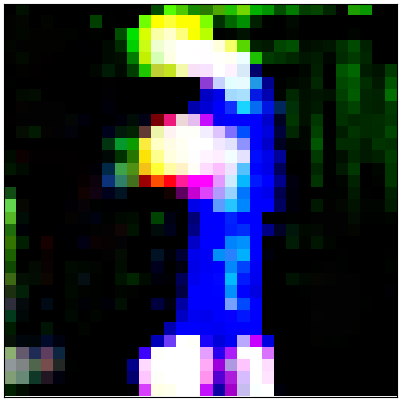}
\label{fig:cifar_32b_pgd_2}}} 
\centering{\subfigure[]{
\includegraphics[width=\cifarwid]{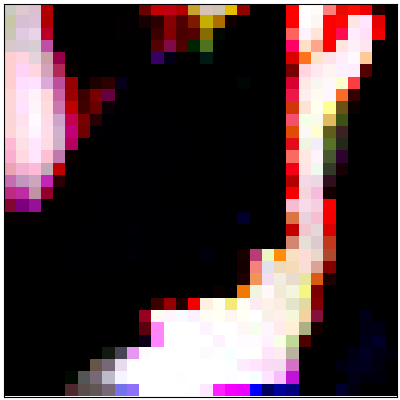}
\label{fig:cifar_32b_pgd_3}}} 
\centering{\subfigure[]{
\includegraphics[width=\cifarwid]{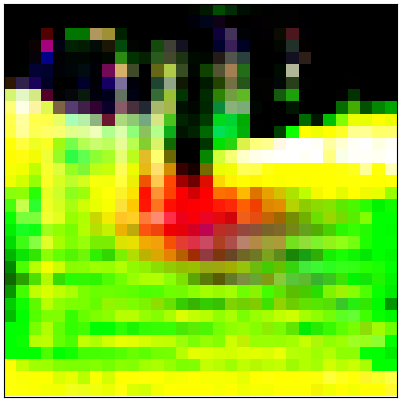}
\label{fig:cifar_32b_pgd_4}}} 
\centering{\subfigure[]{
\includegraphics[width=\cifarwid]{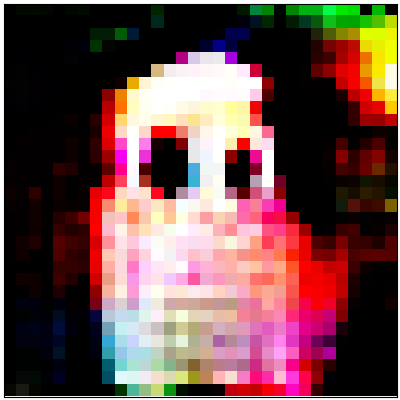}
\label{fig:cifar_32b_pgd_5}}} 
\centering{\subfigure[]{
\includegraphics[width=\cifarwid]{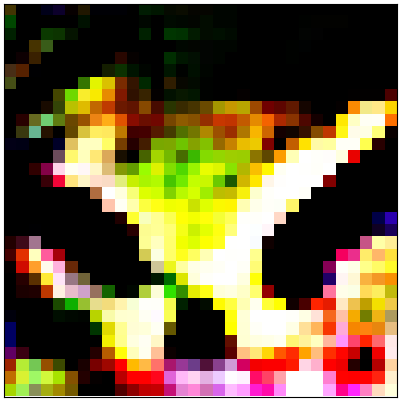}
\label{fig:cifar_32b_pgd_6}}} 
\centering{\subfigure[]{
\includegraphics[width=\cifarwid]{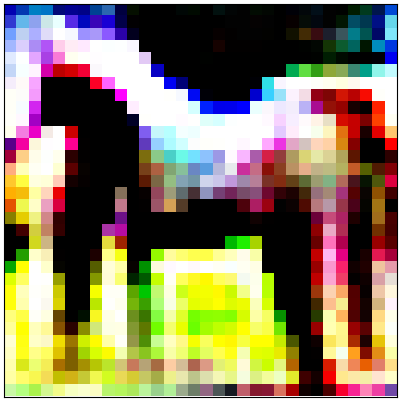}
\label{fig:cifar_32b_pgd_7}}} 
\centering{\subfigure[]{
\includegraphics[width=\cifarwid]{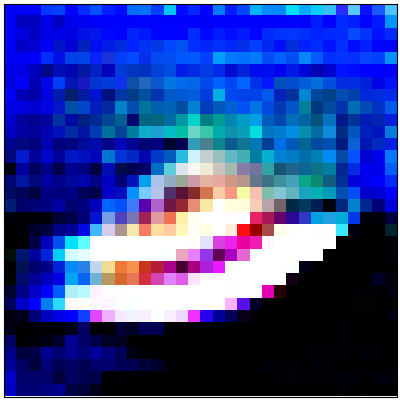}
\label{fig:cifar_32b_pgd_8}}} 
\centering{\subfigure[]{
\includegraphics[width=\cifarwid]{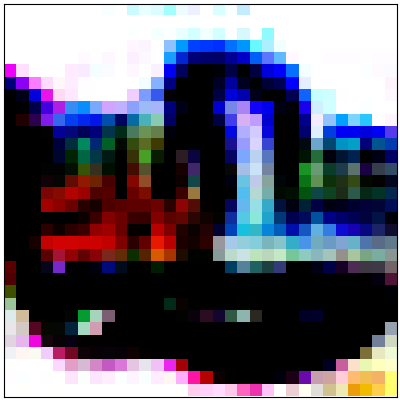}
\label{fig:cifar_32b_pgd_9}}} 
\caption{Generating CIFAR-10 examples with a full-precision discriminative
model trained with 20 iterations of PGD~\cite{madry2018towards} with
$\epsilon=0.3$, and 0.05 L2 weight decay on first and last layers. Gradient
ascent performed on a noisy image for 100 iterations with a step size of
1e-2. Predictions are made with 28.6\%--53.6\% [\subref{fig:cifar_32b_pgd_4} and \subref{fig:cifar_32b_pgd_2}] confidence.}
\label{fig:cifar_32b_pgd_gen}
\end{figure}

\begin{figure}[t!]
\centering
\includegraphics[width=\columnwidth]{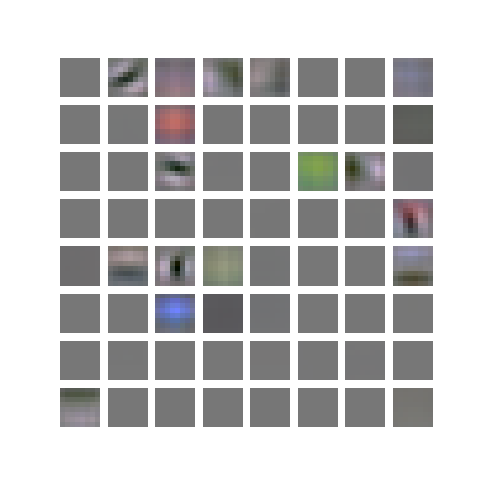}
\vspace{-12mm}
\caption{A visualization of all sixty-four ($nf=64)~8\times8$ convolution 
filters
from the first layer of our three-layer convolutional network trained
on CIFAR-10 with 20 steps of PGD up to $\epsilon_{\max}=0.3$
with an $L_\infty$ norm. A regularization constant of 0.05
was used with $L_2$ weight decay. The model maintains 72\% of its 56\%
clean test set accuracy when attacked with an
$L_\infty$--$\epsilon_{\max}=\nicefrac{8}{255}\approx0.031$ constrained
PGD adversary for 50 attack steps.}
\label{fig:a64_4_conv1_body}
\end{figure}

It is interesting to observe that the $L_2$ weight decay used with the model
visualised in Figure~\ref{fig:a64_4_conv1_body} does not have the effect one
might naively assume: to suppress the magnitude of the~\emph{average} element
across~\emph{all} kernels, more or less equally.
Instead, the weight decay objective is realized by dropping the majority of
filters entirely ($\approx 72\%$), and focusing on primitive concepts. A
dedicated filter for each of the primary colours (R, G, B) is learned, along
with some vertical, diagonal, and horizontal edge detectors. We suspect this is
a near-optimal strategy for a model of this architecture and capacity as these
features clearly generalize to a broad class of naturally occurring images,
while minimizing sensitivity along directions of low variance in the data. This
is reflected by the excellent robustness of the model under a very strong PGD
attack, and it retains a significantly higher percentage of its natural test
accuracy than the state-of-the-art models defended with PGD\@. This model
achieves just 5.6\% under the state-of-the-art robustness of 45.8\% for 
CIFAR-10 for the same PGD attack set on the WideResNet (W32--10)
of~\citet{madry2018towards}, despite having 31.7\% less clean test accuracy and 
only 1.5\% as many parameters. These results concur with 
overwhelming evidence that modern deep neural networks are 
over-parameterized, e.g see~\cite{rosenfeld2018intriguing}.


\subsection{ImageNet Classification}

\begin{table}[t!]
\centering
\caption{Single crop (224px) ResNet-18 error rates vs.~FGSM
perturbation amount for ImageNet misclassification attack for various
settings of precision for weights (W), activations (A), and gradients (G). 
Attacks used 32-bit gradients to minimize~\emph{gradient masking}, regardless
of the gradient precision used during training. ResNets were only trained
for 76 epochs rather than the full 110 in~\citet{zhou2016dorefa} to reduce
overfitting.}
\label{tab:imagenet-resnet}
\begin{tabular}{|c|l|l|l|l|l|} 
\hline 
W,A,G                     & Error & Clean  & eps=2  & eps=4  & eps=8  \\ \hline 

\multirow{2}{*}{32,32,32} & Top-1 & \textbf{38.8\%} & 97.9\% & 98.8\% & 98.8\%
\\
\cline{2-6}
                          & Top-5 & \textbf{16.2\%} & 83.5\% & 88.7\% & 90.7\%
                          \\ \hline 

\multirow{2}{*}{3,3,32}   & Top-1 & 40.7\% & 97.8\% & 98.8\% & 98.9\% \\
\cline{2-6}
                          & Top-5 & 17.3\% & 83.1\% & 88.7\% & \textbf{90.7\%}
                          \\ \hline 

\multirow{2}{*}{3,3,8}    & Top-1 & 40.2\% & 97.8\% & 98.7\% & 98.8\% \\
\cline{2-6}
                          & Top-5 & 17.2\% & 83.5\% & 89.0\% & 90.8\% \\ \hline 

\multirow{2}{*}{3,3,6}    & Top-1 & 40.5\% & 97.6\% & 98.7\% & \textbf{98.7\%}
\\
\cline{2-6}
                          & Top-5 & 17.2\% & \textbf{82.8\%} & \textbf{88.5\%}
                          & \textbf{90.7\%}
                          \\ \hline 

\multirow{2}{*}{2,2,32}   & Top-1 & 45.4\% & 97.4\% & 98.5\% & 98.9\% \\
\cline{2-6}
                          & Top-5 & 20.4\% & 83.3\% & 88.9\% & 91.2\% \\ \hline 

\multirow{2}{*}{2,2,6}    & Top-1 & 44.9\% & 97.5\% & 98.8\% & 98.9\% \\
\cline{2-6}
                          & Top-5 & 20.4\% & 83.3\% & 89.3\% & 91.3\% \\ \hline 

\multirow{2}{*}{1,2,32}   & Top-1 & 50.4\% & 97.9\% & 98.8\% & 99.1\% \\
\cline{2-6}
                          & Top-5 & 24.8\% & 86.9\% & 91.8\% & 93.8\% \\ \hline 

\multirow{2}{*}{1,2,8}    & Top-1 & 51.1\% & 97.8\% & 98.7\% & 98.8\% \\
\cline{2-6}
                          & Top-5 & 25.6\% & 86.7\% & 91.5\% & 93.1\% \\ \hline 

\multirow{2}{*}{1,2,6}    & Top-1 & 50.2\% & 97.9\% & 98.9\% & 98.9\% \\
\cline{2-6}
                          & Top-5 & 24.8\% & 86.8\% & 91.8\% & 93.4\% \\ \hline 

\multirow{2}{*}{1,1,32}   & Top-1 & 59.6\% & \textbf{96.8\%} & \textbf{98.4\%}
& 98.9\%
\\
\cline{2-6}
                          & Top-5 & 33.5\% & 86.4\% & 92.1\% & 94.2\% \\ \hline 

\multirow{2}{*}{1,1,8}    & Top-1 & 59.6\% & 96.9\% & 98.5\% & \textbf{98.7\%}
\\
\cline{2-6}
                          & Top-5 & 33.5\% & 86.4\% & 92.0\% & 93.7\% \\ \hline 

\end{tabular}
\end{table}

The data in Table~\ref{tab:imagenet-resnet} could be interpreted in one 
of two ways, either that ResNets are fragile to precision reduction, or that
they are prone to overfitting since their improvements on the natural test 
set consistently disappear compared to low-precision variants 
when~\emph{attacking} with various perturbation sizes. The latter argument 
implies that the incremental accuracy improvement of the full-precision model 
on the test set does not generalize. We also found that retraining the 
WideResNet from~\citet{madry2018towards} with one or two more orders of 
magnitude $L_2$ weight decay, but less than used with our vanilla 
CNN, fails to yield more than 30\% test accuracy with otherwise identical 
hyper-parameters.

Although the error rates in Table~\ref{tab:imagenet-resnet} are high, none of 
the models were trained with data augmentation and~\emph{no} other defense 
for ImageNet currently exists that does much better than what we show here. 
The defense of~\citet{xie2018mitigating} consisting of a randomization input 
layer was found to achieve 0\% accuracy by~\citet{athalye_obfuscated_2018} for 
the same threat model ($L_\infty$ and $\epsilon=8$), although we acknowledge 
that a stronger iterative attack was used in that case. We suggest that 
starting with smaller models that can handle strong weight decay or
precision reduction is worth exploring as a natural defense.

\vspace{-3mm}

\section{Discussion}

It is important that we move away from data augmentation-based techniques for
conferring robustness. In addition to slowing down training by a
factor equal to the number of steps of gradient ascent per weight
update, these techniques have significant limitations. Models defended this way
are vulnerable to perturbations just a small distance beyond that
used during training, which can be observed in Figure 2 of 
the~\cite{madry2018towards} paper. As illustrated by~\citet{TanayG16}, a
prerequisite for generating high quality adversarial examples during
training is a well regularized model, otherwise, the examples will lie extremely
close to the decision boundary and sub-manifolds. As we showed, an attacker is 
likely to be able to find a different attack that works well on a model 
defended with a particular variant of adversarial training.
Conversely, there is no reason to expect properly regularized models to suffer 
from these same limitations, as they fundamentally alter the geometry of the 
decision boundary, and maximize entropy in the unexplored representation space.

\vspace{-3mm}

\section{Conclusion}
\label{Conclusion}

We have shown that small, regularized models, retain a high percentage 
of their natural test accuracy against adversarial examples.
A promising direction for future research in robust machine learning, 
without relying on data augmentation, is to start small. Once 
it can be shown that meaningful features are learned, e.g.,~by testing that 
whatever little accuracy that is obtained does not degrade with local 
adversarial and global non-examples, then progressively add capacity in an 
iterative loop until satisfactory performance is reached.

\subsubsection*{Acknowledgments}

The authors wish to acknowledge the financial support of NSERC, CFI and
CIFAR\@. The authors also acknowledge hardware support from NVIDIA and
Compute Canada. We thank Brittany Reiche for editing and improving the
readability of our manuscript.

\bibliography{icml2018_conference}
\bibliographystyle{icml2018}

\end{document}